\documentclass{article}

\usepackage{microtype}
\usepackage{graphicx}
\usepackage{subfigure}
\usepackage{subcaption}
\usepackage{booktabs} 
\usepackage{xcolor}
\usepackage[numbers]{natbib}
\usepackage{amsmath}
\usepackage{amssymb}
\usepackage{todonotes}
\usepackage{float}
\usepackage{enumitem}
\usepackage{lineno}

\usepackage{hyperref}

\usepackage{xcolor}
\definecolor{imperialblue}{rgb}{0, 0, 0.804}
\definecolor{crimson}{rgb}{0.8627, 0.078, 0.23529}
\definecolor{springgreen}{rgb}{0, 0.5, 0.5}
\definecolor{violet}{rgb}{0.7803, 0.082, 0.52156}
\definecolor{indigo}{rgb}{0.294, 0.0, 0.5098}

\newcommand{\csection}[1]{\section{\textcolor{imperialblue}{#1}}}
\newcommand{\csubsection}[1]{\subsection{\textcolor{springgreen}{#1}}}
\newcommand{\csubsubsection}[1]{\subsubsection{\textcolor{violet}{#1}}}

\usepackage{titlesec}

\usepackage[accepted]{icml2021}

\begin{document}
\nolinenumbers

\twocolumn[
\icmltitle{\textcolor{black}{Kryptonite-N: Machine Learning Strikes Back}}

\icmlsetsymbol{equal}{*}

\begin{icmlauthorlist}
\icmlauthor{Nathan Bailey*}{Imp}
\icmlauthor{Kira Kim*}{Imp}
\icmlauthor{Albus Li*}{Imp}
\icmlauthor{Will Sumerfield*}{Imp}

\end{icmlauthorlist}

\icmlaffiliation{Imp}{Department of Computing, Imperial College London, London, United Kingdom}

\icmlcorrespondingauthor{Albus Li}{yizhuo.li24@imperial.ac.uk}

\icmlkeywords{Machine Learning, ICML}

\vskip 0.3in
]



\printAffiliationsAndNotice{\icmlEqualContribution} 

\begin{abstract}
    Quinn et al propose challenge datasets in their work called ``Kryptonite-N". These datasets aim to counter the universal function approximation argument of machine learning, breaking the notation that machine learning can ``approximate any continuous function" \cite{original_paper}. Our work refutes this claim and shows that universal function approximations can be applied successfully; the Kryptonite datasets are constructed predictably, allowing logistic regression with sufficient polynomial expansion and L1 regularization to solve for any dimension N.    
\end{abstract}

\csection{Introduction}

    Recent advancements in machine learning, namely LLMs like GPT-4, have gained not only popularity from its success in solving various complex problems \cite{palm} but also some scepticism in its capabilities, as suggested by the ``Kryptonite-N" paper \cite{original_paper}. Quinn and Luther state despite applying binary logistic regression with a GPT-driven basis expansion technique, they still fail to achieve high-quality performance on the datasets. 

    In this paper we refute the claims of the ``Kryptonite-N" paper by showing that several models yield good results on the dataset, therefore proving the validity of \emph{Universal Function Approximation}. At first a comprehensive data exploration is performed to inform the experimental design. Based on the experiments, we discuss insights gained from the results followed by an analysis on sustainability of model lifecycle. Furthermore, we demonstrate \hyperref[sec:gpt]{the flaws in their GPT approaches} and explain observed poor performance. Finally, we outline the process where we \hyperref[sec:secret]{determined the construction of the datasets}.
    
\csection{Data Exploration} \label{sec:data_exp}

Initial data exploration using quantitative metrics and visual representation is crucial to gain insights for data manipulation and modelling steps. We exclusively examined the training portion of the datasets, to ensure that we didn't bias our models.

After assessing metrics from each dimension, we discovered unusual results: each dimension in each K-N dataset has an approximate mean of 0.5, standard deviation of 0.5, and range of $ [ 0 - \epsilon, 1 + \epsilon] $. This prompted us to evaluate the data distribution by plotting the \hyperref[fig:pmf]{Probability Mass Function of each dimension for the K-9 Dataset}.

We learned several very important things from this graph. Firstly, each dimension appears to come from a multimodal distribution with peaks at at 0 and 1. Each peak within a dimension takes the same shape, with there being three different distributions seen across the dimensions, which we classify as "Burst-like", "Gaussian-like", and "Spread-like". Secondly, we see that there appears to be no correlation between which peak a data-point is sampled from and the label. This made us curious about the correlation between the dimensions and the labels, which we hypothesized to be low based on this graph.

A \hyperref[fig:CorrMat]{correlation matrix} shows our hypothesis is correct - there is close to zero correlation across dimensions. This implies the labels must be based on a non-linear combination of features. We visualised the data via a \hyperref[fig:PCA]{2-Dimensional Principal Component Analysis} to better understand the data's relationship with the labels, but that resulted in a circle of data points with little clustering of labels, especially at higher dimensions.

However, if we only visualise as \hyperref[fig:2Dim]{slices in 2D}, we can extrapolate that the points fall in $k^k$ distinct clusters, evenly spaced over $k$-Dimensional space.
This reminded us of the XOR problem, a dataset which Multi-Layer Perceptrons fail to solve without non-linearities. The XOR function has points at each corner, with a non-linear relationship to the labels dependent on each input feature (which are also exactly uncorrelated).
It seemed to us that this dataset is similar to the K-N datasets, and therefore that we might use similar methods to solve them, such as polynomial features and basis expansion. This also leads us to the idea of discretizing the dataset, by mapping $v < 0.5 \Rightarrow 0, v>0.5 \Rightarrow 1 $ for each dimension. This method assumes the distributions of each dimension are inherently noisy, which is backed by the fact that the labels seem to fall evenly across the distribution.

\csection{Methodology}

We tackle the Kryptonite-N dataset in the same way as the original work; a classification task with a labelled dataset \(D = \{(x^{(i)}, y^{(i)})\}_{i=1}^N \). We train a parametric model \(f\) with parameters \(\theta\). The training function will seek to optimise these parameters with respect to a loss function \(L\) and return the model with the parameters producing the smallest loss.

\csubsection{Data Standardisation} \label{sec:data_std}
Standardisation is a common pre-processing step to place attributes in uniform scale. The mean \(\mu\) and standard deviation \(\sigma\) of each feature \(j\) are computed for the dataset and each value is scaled such that each feature has a mean of 0 and a standard deviation of 1. \(\hat{x}_{ij} = \frac{x_{ij} - \mu_{j}} {\sigma_{j}}\)

It is especially important for models using linear combinations of weights and inputs, \(\hat{y} = \theta^Tx\) and optimizing using gradient descent. Consider a single layer neural network, \(\frac{\partial L}{\partial W} = x\frac{\partial L}{\partial z}\) where \(z\) is the output of the layer. Therefore, all partial derivatives are scaled by the input features. If some features are centred around a larger mean, they will cause larger updates to their corresponding weights. Standardising data counters against this better ensures convergence. It also enables relative comparison of polynomial coefficients for polynomial features which will be essential in understanding underlying patterns across feature values.

\csubsection{Neural Networks}
Neural Networks are universal function approximations and therefore are a good solution to experiment with to solve the Kryptonite-N dataset. Given an input vector \(x\), a single layer \(l\) of a neural network will map this to output using a matrix of weights \(W_{l}\) and a vector of biases \(b_{l}\). The number of rows in \(W\) corresponds to the number of neurons in the layer, with each neuron connecting to all inputs with a corresponding weight. Mathematically: \(A_{l} = \sigma_{l}(W_{l}x^T + b_{l}\)). Where \(\sigma_{l}\) is a non-linear activation function of the layer. An activation function is key to introduce non-linearity into the network, allowing it to approximate non-linear functions. 

Layers of a neural network can be stacked to produce a multi-layer network, defined as: \(\hat{y} = \sigma_{L}(W_{L}(\sigma_{L-1}(...\sigma_{0}(W_{0}x^T+b_{0}))+ b_{L})\). Where \(L\) is the final layer in the network. In a multi-layer neural network, there are 3 hyper-parameters to optimize for: the number of layers, the number of neurons in each layer and the activation function used in each layer. 
\vspace*{-0.7\baselineskip}
\csubsubsection{Mini-Batch Gradient Descent} \label{sec:batch}

For a given weight in a neural network, the update rule is given by \(w \leftarrow w - \alpha\nabla_{w}L\).

Stochastic gradient descent will compute the loss and the weight for every sample in the dataset. This can introduce noise into the weight updates, making the training process unstable as every derivative can point in different directions.  Batched gradient descent, on the other hand, computes a loss value after processing every sample in the dataset, making one weight update for every pass of the dataset. This has the disadvantage of slow convergence to optimal, as it needs a full pass of the data for a single update.

Mini-batch gradient descent represents a middle ground, where the loss and derivative are calculated over a subset of the samples in the dataset. This helps to reduce the noise from stochastic gradient descent, whilst speeding up convergence compared to batched gradient descent. The batch size hyperparameter \(B\) can be altered to adjust the trade-off between convergence speed and noise reduction. \cite{Géron_2019}[p118-127]

\csubsubsection{Alternative Optimizers} \label{sec:opt}
Stochastic gradient descent (SGD) with Momentum, can be written as shown in section \ref{appendix:sgd}. This performs a gradient descent update whilst taking into account the previous values of the gradient. As optimization continues in the same direction, larger steps are taken, enabling faster convergence to the optimal solution. Additionally, it allows the gradient descent process to converge to the global minima away from any local minima \cite{Géron_2019}[p352]. This adaptation to SGD can add improvement but comes at the cost of an additional hyper-parameter \(\beta\).

An extension to SGD is adaptive moment estimation (Adam), defined as shown in section \ref{appendix:adam} \cite{Géron_2019}[p356]. Adam combines accumulated momentum \(m\), and that of the square of the gradients \(s\). Applying equations 3 and 4 gives bias-corrected estimates, counteracting the initialisation of these parameters to 0 \cite{adam}. Finally, \(\hat{m}\) and \(\hat{s}\) are applied to \(\theta\). Performing element-wise scaling by \(\hat{s}\) in the final term decays the learning rate proportional to the size of the individual parameter gradients. This encourages a more direct solution to the minimum, rather than following a curve, which SGD with momentum would exhibit \cite{Géron_2019}[p355]. Using Adam introduces 2 additional hyper-parameters \(\beta_{1}\) and \(\beta_{2}\).

\csubsubsection{Dropout}
Dropout is used to counteract overfitting in neural networks. Mathematically, a neuron in a layer has probability \(p\) to output a zero (and be dropped out) in a single forward pass. For an input \(x\), we have \(\hat{x} = r * x\), where \(r \sim Bernoulli(p)\) and \(\hat{x}\) becomes the input into the layer \cite{dropout}. Dropout creates a new sampled network at each forward pass, encouraging neurons to use all of their inputs, rather than just focusing on a subset. This creates a network that exhibits better generalization capabilities \cite{Géron_2019}[p365-366] \cite{Foster2019}[p50]. For each layer \(l\) that uses dropout, we have an additional hyper-parameter \(p_{l}\) to tune.

\csubsection{Polynomial Basis Expansion Logistic Regression}
\csubsubsection{Basis Expansion}
Basis Expansion increases the dimension of original feature space using non-linear transformation. In particular, Polynomial Basis Expansion introduces polynomial combinations of features: for a given input feature of length $m$ $\mathsf{x} = (x_1, x_2,...,x_m)$, polynomial basis expansion of degree $n$ will generate all polynomials of powers up to degree $n$. Not only does it capture multiple powers of the same feature but also interactive features; and products of distinct features ($x_i, x_j, x_k; i \neq j \neq k$).\cite{scikit-learn} It is particularly useful as the resulting feature space is linearly separable\cite{bishop06} while capturing complex relationships across feature values. 

Defining as a basis function: let $\space \mathsf{x} = (x_1,x_2,..,x_m) \in \mathbb{R}^m$  $\phi(\mathsf{x})$ which will generate following polynomial basis where $\mathsf{x}=(x_1,x_2,x_3), m=3$:
\vspace{-1em}
\begin{center}
    \(\phi: \mathbb{R}^m \rightarrow \mathbb{R}^{\binom{n+m}{n}-1}\)

\(\phi(\mathsf{x}) = [x_1, x_2, x_3, \textcolor{blue}{x_1x_2, x_1x_3, x_2x_3}, x_1^2, x_2^2, x_3^2]^{\top}\)
\end{center}


The function will map to a new feature space with the following basis where interactions coloured in blue. 
\vspace{-0.3em}

\csubsubsection{Logistic Regression}
Logistic Regression is one form of generalised linear model that is suitable for a classification problem like the task in hand, formulated as a composition of $\sigma \circ \phi$ where activation function $\sigma$ is a logistic sigmoid; generating Bernoulli Distributed posterior probabilities $P(y|\mathsf{x}, \mathsf{w})$.\cite{murphy2012,bishop06}

\(y = \sigma(f(x)), \space f(x) = \theta^{\top}\phi(x)\); \(\sigma(f) = \frac{1}{1+\exp{(-f)}} \in [0,1]\)


\csubsubsection{Regularization} \label{sec:reg}
Regularization is used to prevent models from overfitting. L2 regularization adds the squared L2 norm of the weights \(\frac{\lambda}{2}\lVert \textbf{w} \rVert_{2}^2\) as a penalty to the loss function, whereas L1 regularization adds the L1-Norm of the weights \(\frac{\lambda}{2}\lVert \textbf{w} \rVert_{1}\) to the loss function. \(\lambda\) acts as a hyperparameter to control the influence of the penalty on the loss function.

Both L1 and L2 regularization aim to constrain the model to reduce overfitting. L2 encourages the individual weights to take on smaller values, whereas L1 regularization encourages sparsity of the weight vector \(\textbf{w}\).

\csection{Experimental Design}

Given that there is an underlying pattern in the dataset, as specified in the original work, we hypothesise that the Kryptonite-N can be solved using a universal function approximation tool.

\csubsection{Neural Networks} \label{sec:nn_design}

Training a neural network, was sufficient to achieve the desired accuracy on Kryptonite datasets N9-18, with the results shown in the subsequent section. 

For all datasets, the data was first scaled as detailed in section \ref{sec:data_std} and prepared with a training/validation/testing split of 60\%, 20\%, and 20\%. Validation data was used to manually tune hyperparameters, whilst the test data was used as a hold-out test set to measure the generalization capabilities of the trained model. Before splitting, the data was shuffled with a random seed to ensure a diversity of samples across the three sets. 

To reduce the amount of overfitting, a simpler neural network was favoured over a more complex one. Therefore, we kept the number of hidden layers to 1, which used 72 neurons. This network architecture selected was performant and generalized to all N9-18 datasets. Either a Tanh or an ELU activation function was used in all the penultimate layers, with a sigmoid activation in the final layer to output $p \in [0,1]$ (all activation functions are detailed in section \ref{appendix:act}), interpreted as a probability. Dropout was used for some networks between the hidden and output layers, to control overfitting.

Each neural network was trained using either an SGD with momentum or an Adam optimizer, as we detail in section \ref{sec:opt}. The networks were trained with batched data for 1000 epochs, with early stopping \cite{Prechelt2012} employed to stop after 200 epochs with no improvement on the validation data. The network with the best validation loss (binary cross-entropy loss - shown in section \ref{appendix:bce}) was used as the final trained network.

\csubsection{Polynomial Basis Expansion Logistic Regression} \label{sec:poly}
The first iteration was following the same implementation of the original work but with standardised dataset. Standardisation of input features greatly improved the training accuracy for N=9 albeit with some overfitting and small improvements for larger N.

In order to understand the effect of including all polynomial combinations up to degree $N$, it was necessary to assess the relative contributions of each polynomial feature towards the outcome by analysing the standardised coefficients. Notably, for N=9, weights $\boldsymbol{\theta}$ were sparse with the majority close to 0 with one feature $x_1x_3x_5x_7x_8$ being the key driver. This prompted next iteration to include interaction features only.

\texttt{GridSearchCV} is used to test various powers of polynomials and optimisation method other than the default used in \texttt{LogisticRegression}. Logistic Regression with SGD was tested using \texttt{SGDClassifier}. The experiment was iterative from assessing accuracy values with the best combination of hyperparameter values, evaluating standardised polynomial coefficients to validate strong contributors, rerunning the search with a modified range of hyperparameters.   

Data fitting over varying degrees of polynomials is evaluated using boxplots and trajectories of the loss function in cross-validation. \textit{StratifiedKFold} Cross Validation is used to assess mean accuracy on test sets keeping label distribution consistent across samples.

\csection{Experimental Results} \label{sec:results}
\csubsection{Neural Networks}
The architectures of each neural network used for each dataset are shown in table \ref{table:nn} within the appendices. As mentioned in section \ref{sec:nn_design}, the same baseline network was used, a single-layer network with 72 neurons. Across the datasets, we varied the learning rate, optimiser, activation function, batch size and dropout rate.
\vspace*{-0.5\baselineskip}

\csubsubsection{Tuning Hyperparameters}
For hyperparameter tuning, a baseline network comprised of 72 neurons, ELU activation function and SGD with Momentum was used. Using the validation dataset, optimiser, learning rate, batch size and dropout probability \(p\) were fine-tuned manually to achieve the desired accuracy on the test set as reported in the original work. Due to the vast amount of hyperparameter combinations, performing an exhaustive sweep of the hyperparameter space in the given time was not feasible. However, ideally, this should be explored in the future, as additional performance could be gained.

Across all datasets, we noticed that a larger batch size generally led to increased performance on the validation dataset.
For the N-9 dataset, the baseline network was adequate to reach the desired test accuracy with no changes applied. 

For the N-12 dataset, the baseline network exhibited small overfitting and was countered by adding dropout after the hidden layer with \(p=0.01\). This helped reduce overfitting and reach the desired test accuracy. 

For the N-15 dataset, we found replacing the ELU activation layer with a Tanh layer yielded performance gains. In addition, as in the N-12 dataset, an increase in the number of input features led to increased overfitting, which was tackled by increasing the dropout rate to $p=0.03$.

Lastly, for the N-18 dataset, the same Tanh activation function was used as in N-15. However, we saw large performance gains when using the Adam optimizer over SGD with momentum. This dataset also followed the trend of an increase in overfitting due to the increased number of features. The dropout parameter was changed to $p=0.07$.

For all networks, the improvements gained from tuning hyperparameters are shown in figures \ref{fig:overfit_N12} to \ref{fig:overfit_N18} within the appendices.

\csubsubsection{Neural Network Results}

The results for the tuned neural networks on the test data are shown in table \ref{table:nn} where we report average test accuracy with standard deviation after 10 training and testing runs for each network - where each reaches the target accuracy.

The training curves for each neural network are detailed in figures \ref{fig:neural_net_results_N9} to \ref{fig:neural_net_results_N18} in the Appendix. Across all datasets, the selected models show little to no overfitting with slight deviations from the mean. This shows our networks exhibit strong stability in training and generalisation to new unseen data. Due to this, we confidently expect to see similar accuracy on new data generated from the same underlying distribution. 

\vspace*{-1.3\baselineskip}
\begin{table}[h]
\caption{Neural Networks Results}
\label{table:nn}
\vskip 0.15in
\begin{center}
\begin{small}
\begin{sc}
\begin{tabular}{ccc}
\toprule
Dataset    & Target Accuracy   & Test Accuracy   \\
\midrule
N-9        & 0.95                  &   \(0.9553 \pm 0.0011\)           \\
N-12       & 0.925                  &   \(0.9553 \pm 0.0090\)               \\
N-15       & 0.9                  &     \(0.9271 \pm 0.0051\)             \\
N-18       & 0.875                  &      \(0.9068 \pm 0.0110\)            \\
\bottomrule
\end{tabular}
\end{sc}
\end{small}
\end{center}
\vskip -0.1in
\end{table}
\vspace*{-1.3\baselineskip}

\csubsection{Polynomial Basis Expansion Logistic Regression}
The model performed best with \textit{SGD} optimisation with L1 regularisation taking \textit{standardised, interaction only} polynomial expanded basis as inputs \textit{using only a limited degree powers} of $(\frac{2}{3}*N,N)$ (accuracy distribution evaluated over multiple models in Appendix \ref{appendix:poly_basis_plot}) for N=9, 12 and 15 with exact parameters outlined in Table \ref{table:lg_hyp}. Stricter regularisation was necessary for greater N to assure convergence with better generalisation as we encountered a sparse weight landscape. The higher degrees of the polynomial were performant with the largest contribution signals coming from interaction features with a certain proportion of feature values.

Table \ref{table:lg_score} summarises the mean and standard deviation of cross-validated scores of 10 splits, from each fitted model with the above hyperparameters for varying $N$. This approach displayed sub-optimal performance for N greater than 15 suggestive of different variations of polynomial features may be required that prompted additional experiments detailed in Section \ref{sec:secret}.

\vspace*{-1.3\baselineskip}
\begin{table}[H]
\caption{Hyperparameters for Basis Expansion \& Logistic Regression}
\label{table:lg_hyp}
\vskip 0.15in
\begin{center}
\begin{small}
\begin{sc}
\begin{tabular}{cccc}
\toprule
Dataset    & Degree   & Regularisation & $\lambda$ value \\
\midrule
N-9        & (5,7)   & l2  & 0.01\\
N-12       & (8,10)  &  l2 & 0.01\\
N-15       & (10,14) &  l1 & 0.015\\
\bottomrule
\end{tabular}
\end{sc}
\end{small}
\end{center}
\vskip -0.1in
\end{table}

\vspace*{-1\baselineskip}

\begin{table}[h]
\caption{Logistic Regression Results}
\label{table:lg_score}
\vskip 0.15in
\begin{center}
\begin{small}
\begin{sc}
\begin{tabular}{cccc}
\toprule
Dataset    & Target Accuracy   & Test Accuracy   \\
\midrule
N-9        & 0.95              &   \(0.9582 \pm 0.0047\) \\
N-12       & 0.925             &   \(0.9638 \pm 0.0032\)  \\
N-15       & 0.9               &     \(0.9572 \pm 0.0079\) \\
\bottomrule
\end{tabular}
\end{sc}
\end{small}
\end{center}
\vskip -0.1in
\end{table}

\vspace*{-0.5\baselineskip}
\csection{Dataset Secrets Revealed and Further Logistic Regression}
\label{sec:secret}
Through a series of well-thought-out experiments and coincidental discoveries, we successfully unveiled the secrets of curating the Kryptonite-$N$ dataset, which is a \emph{high-dimensional XOR problem with 1/3 redundant features}. We revealed the secrets of the dataset through a series of well-thought-out experimental methodology. The detailed steps of our discovery process can be found in Appendix \ref{appendix:steps}, which we \textbf{strongly recommend reading}. This section outlines the mathematical formulation of the dataset at first. The experimental results of our best model using Logistic Regression with L1-Regularization and an optimal logistic regression model with feature filtering oracle are also provided.

\csubsection{Mathematical Formulation of the Dataset}
The Kryptonite-$N$ dataset is characterized by a feature space where part of inputs is unrelated to the final output - such input will be referred to as \emph{irrelevant features}. While its counterparts, which directly influence the final output, will be termed \emph{informative features}. Notably, in Kryptonite-$N$ Dataset, exactly \(\frac{1}{3}N\) features are redundant.
Specifically, feature index space is defined as 
\vspace{-.2cm}
\[
F = \{1, 2, \dots, N\}, \quad |F| = N 
\]\\[-.55cm]
\emph{Informative features} are represented by the index space\\[-.3cm]
\[
F_{\textit{info}} = \{a_i\}, \quad F_{\textit{info}} \subset F, \quad |F_{\textit{info}}| = \frac{2}{3}N
\]\\[-.4cm]
Conversely, \emph{irrelevant features} are represented by the complement set \\[-.4cm]
\[
F_{\textit{irre}} = \overline{F_{\textit{info}}}, \quad |F_{\textit{irre}}| = \frac{1}{3}N
\]\\[-1cm]

After filtering informative features (the method to be detailed in the proceeding section), we perform a discretization of the data. According to \ref{sec:data_std} each feature displays a binomial distribution trend, and the discretized value is based on which part it belongs to. 
Specifically, feature $x_i$ after discretization is: \(x_i^{dis} = \mathbf{1}_{\{x_i > \overline{X_i}\}}\)

where \(\overline{X_i}\) represents feature $i$'s mean value across dataset.

The dataset is fundamentally a high-dimensional XOR problem, where the target label $y$ is derived as: \(y = \bigoplus_{i\in F_{\textit{info}}}x_i^{dis}\) where \(\bigoplus\) is the summation notation for XOR operator.

The XOR structure ensures non-linearity, making simple models struggle to learn effectively without appropriate basis expansion.

\csubsection{Testing Results}

\setlist[enumerate]{leftmargin=7pt}
\setlist[itemize]{leftmargin=7pt}

\begin{enumerate}
    \item \textbf{XOR Conjecture Verification}\\
    As a verification of our conjecture, the second column of Table \ref{table:xor} presents the test accuracy on the entire dataset using our handcrafted feature selection, discretization, and XOR operator functions (XOR).
    \vspace{-.5em}
    \item \textbf{Logistic Regression with L1 Regularisation}\\
    Logistic Regression used here is \emph{Gradient-based} using \texttt{SGDClassifier} in \texttt{scikit-learn}, as discussed in Appendix \ref{appendix:steps}. Only results of Logistic Regression with L1 Regularisation (LR with L1) on $N$ = 9, 12, 15, 18 are displayed in the third column of Table \ref{table:xor} since we can't do polynomial basis expansion on larger $N$ with limited RAM space. However, it should achieve the same good performance on high-dimensional space theoretically.
    \vspace{-.5em}
    \item \textbf{Logistic Regression with Feature Selection Oracle}\\
    With Feature Selection Oracle (FSO) based on feature distribution shape we observed in Step 6 of Appendix \ref{appendix:steps}, we reduced the feature space to 1 dimension only including the multiplication of \emph{informative features}, formulated as follows: \(input = \prod_{i\in F_{\textit{info}}}x_i^{norm}\).
    Where \(norm\) implies normalised. Experimental results of Logistic Regression under such conditions (LR with FSO) can be seen in the fourth column of Table \ref{table:xor}.
\end{enumerate}

\vspace*{-0.8\baselineskip}
\begin{table}[h]
\caption{Final Accuracy Results}
\label{table:xor}
\vskip 0.05in
\begin{center}
\begin{small}
\begin{sc}
\begin{tabular}{cccc}
\toprule
Dataset & XOR & LR with L1 & LR with FSO  \\ 
\midrule
N-9        &   \(0.9583\)   & \(0.9599\pm0.0040\) & \(0.9577\pm0.0075\) \\
N-12       &   \(0.9642\)   & \(0.9657\pm0.0032\) & \(0.9663\pm0.0037\) \\
N-15       &     \(0.9672\) & \(0.9697\pm0.0026\) & \(0.9680\pm0.0036\) \\
N-18       &     \(0.9710\) & \(0.9708\pm0.0013\) & \(0.9719\pm0.0029\) \\
N-24       &     \(0.9751\) & \(N/A\) & \(0.9748\pm0.0018\) \\
N-30       &     \(0.9764\) & \(N/A\) & \(0.9755\pm0.0016\) \\    
N-45       &     \(0.9805\) & \(N/A\) & \(0.9805\pm0.0015\) \\    
\bottomrule
\end{tabular}
\end{sc}
\end{small}
\end{center}
\vskip -0.1in
\end{table}

\csection{Discussion}
\csubsection{Model Performance Insights}
The experimental results highlighted different behaviours for logistic regression (with polynomial basis expansion) and neural networks, which provided insights on various aspects of fundamental problems in Machine Learning:
\vspace{-0.5em}
\begin{enumerate}
\itemsep0em 
    \item \textbf{Underperforming Neural Networks in High Dimensions}
    Neural networks struggled for Higher N due to the sparsity and combinatorial nature of the dataset. High-dimensional spaces introduce the ``curse of dimensionality" \cite{bellman1961}, where the volume of the feature space increases exponentially, making it harder to generalize from sparse data. This was apparent in the models' optimization process facing difficulties in finding the global minimum efficiently \cite{goodfellow2016}.
    
    \item \textbf{Better performance of Logistic Regression}
    Logistic regression performed well due to its linearity and simplicity, focusing on the most informative features. Occam’s Razor, a principle advocating for simpler models with fewer parameters when multiple models explain the data equally well \cite{jaynes2003}, suggests that logistic regression, with feature selection, can be a more effective solution to high-dimensional problems compared to more complex models.

    \item \textbf{Trade-offs in L1/L2 Regularization}
    \vspace{-0.5em}
    \begin{itemize}
    \itemsep0em 
        \item L2 regularization (Ridge regression) smooths the model by penalizing larger weights, which helps prevent overfitting but may obscure important features, as it doesn't null out coefficients. This results in a more general yet less interpretable model \cite{hoerl1970}.
        \item L1 regularization (Lasso regression) encourages sparsity, effectively selecting a subset of features. However, it may eliminate useful features when there are many correlated features, leading to potential underfitting \cite{tibshirani1996}.
    \end{itemize}
    \vspace{-0.5em}
    However, under this specific setting where one feature deterministically outweighs others, L1 proves to be the ideal choice.

    \item \textbf{Insights into XOR Problem and Dataset Design}
    The XOR problem is inherently non-linear, making it difficult for linear models to solve, especially in high dimensions. Besides, Kryptonite-N dataset also highlights how structured redundancy can reveal a model's weaknesses. By embedding these two difficulties, this dataset challenges the \emph{Universal Approximation Theorem} \cite{kolmogorov1965}. It demonstrates that irrelevant features can impair a model's ability to focus on important signals, emphasizing the need for effective feature selection in high-dimensional spaces.
\end{enumerate}


\csubsection{Sustainability Analysis}
Sustainability within machine learning is becoming increasingly more prevalent due to the rise of LLMs. To gauge the carbon impact of our work, we use the common tool "CodeCarbon" to estimate carbon emissions of a Python program \cite{benoit_courty_2024_14062504}. We separated the estimates of carbon footprint for training and inference, where the former expects to yield a higher value as a more resource-intensive process, the latter is equally important to reflect the expected footprint post-deployment. \cite{Luccioni_2024}



Figure \ref{fig:carbon_impact} in the appendix shows the estimated emissions (in C02 eq) for both training (over 10 runs) and inference (batch size of 128). Measurements were performed on a 13-inch MacBook Air with an Apple M3 SoC and models running only on the CPU. Our largest emission, 0.00014kg, is roughly equivalent to 1 metre of driving \cite{c02_conv}. Overall training emissions are low but grow as the number of features increases. Logically, this is sound as a larger network will be needed to process more features. For testing, we should expect the same pattern as in training, and we reason that this could be due to insignificant emissions produced. Nevertheless, if these networks were deployed, careful consideration is needed of the minimum number of features to perform the task. 

One reason why the emissions here are so low is due to the arm architecture of the M3 chip, which uses less power than an x86 counterpart. To gain an insight into the emissions of training and testing these models on different devices, we show graphs of emissions and energy usage for an x86 system in appendix \ref{appendix:carbon}. We see nearly a 3x increase in total energy consumption when switching platforms, highlighting the importance of system choice on emissions when training and testing models.

Whilst gaining knowledge of the carbon footprint of the codebase is important, it does not give the full sustainability picture. Embodied emissions are often overlooked in sustainability analysis as a whole. These are the emissions coming from the manufacturing of the hardware used to train and test the models \cite{faiz2024llmcarbonmodelingendtoendcarbon}. Whilst overlooked, they often contribute a significant amount of total carbon. For example, Hugging Faces' BLOOM LLM has a total carbon footprint of 50 tonnes of C02 eq, with embodied emissions making up 22\% of this \cite{luccioni2022estimatingcarbonfootprintbloom}. Although there are emerging tools to estimate embodied emissions \cite{faiz2024llmcarbonmodelingendtoendcarbon}, generally they are much harder to estimate since the information needed is hidden from the end user. However, we can estimate that a 15-inch MacBook Air with an Apple M3 SoC has a 158kg C02 eq footprint \cite{apple}. The use of the device vastly extends the scope of this work, so we should only contribute a fraction of this to our overall emissions. However, it is useful to understand the vast emissions produced during the manufacturing process. 

Socially and ethically, it is hard to make a judgement on the impact of this work, since we do not know the origins of the dataset. However, proper care of data from a privacy and security perspective must be taken into account if these models are deployed. 

\newpage
\bibliography{main}

\begin{thebibliography}{28}
\providecommand{\natexlab}[1]{#1}
\providecommand{\url}[1]{\texttt{#1}}
\expandafter\ifx\csname urlstyle\endcsname\relax
  \providecommand{\doi}[1]{doi: #1}\else
  \providecommand{\doi}{doi: \begingroup \urlstyle{rm}\Url}\fi

\bibitem[{Apple}()]{apple}
{Apple}.
\newblock {\emph{Product Environmental Report, MacBook Air with M3 chip}. \url{https://www.apple.com/environment/pdf/products/notebooks/M3_MacBook_Air_PER_March2024.pdf}}.
\newblock [Accessed 14 Nov 2024].

\bibitem[Bellman(1961)]{bellman1961}
Bellman, R.
\newblock \emph{Adaptive Control Processes: A Guided Tour}.
\newblock Princeton University Press, 1961.

\bibitem[Bishop(2006 - 2006)]{bishop06}
Bishop, C.~M.
\newblock \emph{Pattern recognition and machine learning}.
\newblock Information science and statistics. Springer, New York, NY, 2006 - 2006.
\newblock ISBN 9780387310732.

\bibitem[Brown et~al.(2020)Brown, Mann, Ryder, Subbiah, Kaplan, Dhariwal, Neelakantan, Shyam, Sastry, Askell, Agarwal, Herbert-Voss, Krueger, Henighan, Child, Ramesh, Ziegler, Wu, Winter, Hesse, Chen, Sigler, Litwin, Gray, Chess, Clark, Berner, McCandlish, Radford, Sutskever, and Amodei]{gpt3}
Brown, T., Mann, B., Ryder, N., Subbiah, M., Kaplan, J.~D., Dhariwal, P., Neelakantan, A., Shyam, P., Sastry, G., Askell, A., Agarwal, S., Herbert-Voss, A., Krueger, G., Henighan, T., Child, R., Ramesh, A., Ziegler, D., Wu, J., Winter, C., Hesse, C., Chen, M., Sigler, E., Litwin, M., Gray, S., Chess, B., Clark, J., Berner, C., McCandlish, S., Radford, A., Sutskever, I., and Amodei, D.
\newblock Language models are few-shot learners.
\newblock In Larochelle, H., Ranzato, M., Hadsell, R., Balcan, M., and Lin, H. (eds.), \emph{Advances in Neural Information Processing Systems}, volume~33, pp.\  1877--1901. Curran Associates, Inc., 2020.
\newblock URL \url{https://proceedings.neurips.cc/paper_files/paper/2020/file/1457c0d6bfcb4967418bfb8ac142f64a-Paper.pdf}.

\bibitem[Chowdhery et~al.(2022)Chowdhery, Narang, Devlin, Bosma, Mishra, Roberts, Barham, Chung, Sutton, Gehrmann, Schuh, Shi, Tsvyashchenko, Maynez, Rao, Barnes, Tay, Shazeer, Prabhakaran, Reif, Du, Hutchinson, Pope, Bradbury, Austin, Isard, Gur-Ari, Yin, Duke, Levskaya, Ghemawat, Dev, Michalewski, Garcia, Misra, Robinson, Fedus, Zhou, Ippolito, Luan, Lim, Zoph, Spiridonov, Sepassi, Dohan, Agrawal, Omernick, Dai, Pillai, Pellat, Lewkowycz, Moreira, Child, Polozov, Lee, Zhou, Wang, Saeta, Diaz, Firat, Catasta, Wei, Meier-Hellstern, Eck, Dean, Petrov, and Fiedel]{palm}
Chowdhery, A., Narang, S., Devlin, J., Bosma, M., Mishra, G., Roberts, A., Barham, P., Chung, H.~W., Sutton, C., Gehrmann, S., Schuh, P., Shi, K., Tsvyashchenko, S., Maynez, J., Rao, A., Barnes, P., Tay, Y., Shazeer, N., Prabhakaran, V., Reif, E., Du, N., Hutchinson, B., Pope, R., Bradbury, J., Austin, J., Isard, M., Gur-Ari, G., Yin, P., Duke, T., Levskaya, A., Ghemawat, S., Dev, S., Michalewski, H., Garcia, X., Misra, V., Robinson, K., Fedus, L., Zhou, D., Ippolito, D., Luan, D., Lim, H., Zoph, B., Spiridonov, A., Sepassi, R., Dohan, D., Agrawal, S., Omernick, M., Dai, A.~M., Pillai, T.~S., Pellat, M., Lewkowycz, A., Moreira, E., Child, R., Polozov, O., Lee, K., Zhou, Z., Wang, X., Saeta, B., Diaz, M., Firat, O., Catasta, M., Wei, J., Meier-Hellstern, K., Eck, D., Dean, J., Petrov, S., and Fiedel, N.
\newblock Palm: Scaling language modeling with pathways, 2022.
\newblock URL \url{https://arxiv.org/abs/2204.02311}.

\bibitem[Courty et~al.(2024)Courty, Schmidt, Goyal-Kamal, MarionCoutarel, Blanche, Feld, inimaz, Lecourt, LiamConnell, SabAmine, supatomic, Léval, LLORET, Cruveiller, Saboni, ouminasara, Zhao, Joshi, Bauer, Bogroff, de~Lavoreille, Laskaris, Phiev, Abati, rosekelly6400, Blank, Wang, Otávio, and Catovic]{benoit_courty_2024_14062504}
Courty, B., Schmidt, V., Goyal-Kamal, MarionCoutarel, Blanche, L., Feld, B., inimaz, Lecourt, J., LiamConnell, SabAmine, supatomic, Léval, M., LLORET, P., Cruveiller, A., Saboni, A., ouminasara, Zhao, F., Joshi, A., Bauer, C., Bogroff, A., de~Lavoreille, H., Laskaris, N., Phiev, A., Abati, E., rosekelly6400, Blank, D., Wang, Z., Otávio, L., and Catovic, A.
\newblock mlco2/codecarbon: v2.7.4, November 2024.
\newblock URL \url{https://doi.org/10.5281/zenodo.14062504}.

\bibitem[Faiz et~al.(2024)Faiz, Kaneda, Wang, Osi, Sharma, Chen, and Jiang]{faiz2024llmcarbonmodelingendtoendcarbon}
Faiz, A., Kaneda, S., Wang, R., Osi, R., Sharma, P., Chen, F., and Jiang, L.
\newblock Llmcarbon: Modeling the end-to-end carbon footprint of large language models, 2024.
\newblock URL \url{https://arxiv.org/abs/2309.14393}.

\bibitem[Foster(2019)]{Foster2019}
Foster, D.
\newblock \emph{Generative deep learning : teaching machines to paint, write, compose, and play}.
\newblock 2019.

\bibitem[Goodfellow et~al.(2016)Goodfellow, Bengio, and Courville]{goodfellow2016}
Goodfellow, I., Bengio, Y., and Courville, A.
\newblock \emph{Deep Learning}.
\newblock MIT Press, 2016.

\bibitem[Géron(2019)]{Géron_2019}
Géron, A.
\newblock \emph{Hands-on machine learning with scikit-learn, Keras, and tensorflow: Concepts, tools, and techniques to build Intelligent Systems}.
\newblock O’Reilly Media, 2019.

\bibitem[Hoerl \& Kennard(1970)Hoerl and Kennard]{hoerl1970}
Hoerl, A.~E. and Kennard, R.~W.
\newblock Ridge regression: Biased estimation for nonorthogonal problems.
\newblock \emph{Technometrics}, 12\penalty0 (1):\penalty0 55--67, 1970.

\bibitem[Jaynes(2003)]{jaynes2003}
Jaynes, E.~T.
\newblock \emph{Probability Theory: The Logic of Science}.
\newblock Cambridge University Press, 2003.

\bibitem[Kingma \& Ba(2017)Kingma and Ba]{adam}
Kingma, D.~P. and Ba, J.
\newblock Adam: A method for stochastic optimization, 2017.
\newblock URL \url{https://arxiv.org/abs/1412.6980}.

\bibitem[Kolmogorov(1965)]{kolmogorov1965}
Kolmogorov, A.
\newblock Three approaches to the quantitative definition of information.
\newblock \emph{International Journal of Computer Mathematics}, 1\penalty0 (1):\penalty0 3--11, 1965.

\bibitem[Luccioni et~al.(2022)Luccioni, Viguier, and Ligozat]{luccioni2022estimatingcarbonfootprintbloom}
Luccioni, A.~S., Viguier, S., and Ligozat, A.-L.
\newblock Estimating the carbon footprint of bloom, a 176b parameter language model, 2022.
\newblock URL \url{https://arxiv.org/abs/2211.02001}.

\bibitem[Luccioni et~al.(2024)Luccioni, Jernite, and Strubell]{Luccioni_2024}
Luccioni, S., Jernite, Y., and Strubell, E.
\newblock Power hungry processing: Watts driving the cost of ai deployment?
\newblock In \emph{The 2024 ACM Conference on Fairness, Accountability, and Transparency}, FAccT ’24. ACM, June 2024.
\newblock \doi{10.1145/3630106.3658542}.
\newblock URL \url{http://dx.doi.org/10.1145/3630106.3658542}.

\bibitem[Minsky \& Papert(1988)Minsky and Papert]{minsky1988perceptrons}
Minsky, M. and Papert, S.
\newblock \emph{Perceptrons: An Introduction to Computational Geometry}.
\newblock MIT Press, expanded edition edition, 1988.

\bibitem[Murphy(2012 - 2012)]{murphy2012}
Murphy, K.~P.
\newblock \emph{Machine learning : a probabilistic perspective}.
\newblock Adaptive computation and machine learning. The MIT Press, Cambridge, Massachusetts, 2012 - 2012.
\newblock ISBN 9780262018029.

\bibitem[{openco2.net}()]{c02_conv}
{openco2.net}.
\newblock {\emph{C02-Converter}. \url{https://www.openco2.net/en/co2-converter}}.
\newblock [Accessed 14 Nov 2024].

\bibitem[Pedregosa et~al.(2011)Pedregosa, Varoquaux, Gramfort, Michel, Thirion, Grisel, Blondel, Prettenhofer, Weiss, Dubourg, Vanderplas, Passos, Cournapeau, Brucher, Perrot, and Duchesnay]{scikit-learn}
Pedregosa, F., Varoquaux, G., Gramfort, A., Michel, V., Thirion, B., Grisel, O., Blondel, M., Prettenhofer, P., Weiss, R., Dubourg, V., Vanderplas, J., Passos, A., Cournapeau, D., Brucher, M., Perrot, M., and Duchesnay, E.
\newblock Scikit-learn: Machine learning in {P}ython.
\newblock \emph{Journal of Machine Learning Research}, 12:\penalty0 2825--2830, 2011.

\bibitem[Prechelt(2012)]{Prechelt2012}
Prechelt, L.
\newblock \emph{Early Stopping --- But When?}, pp.\  53--67.
\newblock Springer Berlin Heidelberg, Berlin, Heidelberg, 2012.
\newblock \doi{10.1007/978-3-642-35289-8_5}.
\newblock URL \url{https://doi.org/10.1007/978-3-642-35289-8_5}.

\bibitem[Quinn \& Luther(2024)Quinn and Luther]{original_paper}
Quinn, H. and Luther, L.
\newblock Kryptonite-n: A simple end to machine learning hype?
\newblock \emph{ICML 2024}, 2024.

\bibitem[Radford \& Narasimhan(2018)Radford and Narasimhan]{gpt}
Radford, A. and Narasimhan, K.
\newblock Improving language understanding by generative pre-training.
\newblock 2018.
\newblock URL \url{https://api.semanticscholar.org/CorpusID:49313245}.

\bibitem[Radford et~al.(2019)Radford, Wu, Child, Luan, Amodei, and Sutskever]{gpt2}
Radford, A., Wu, J., Child, R., Luan, D., Amodei, D., and Sutskever, I.
\newblock Language models are unsupervised multitask learners.
\newblock 2019.
\newblock URL \url{https://api.semanticscholar.org/CorpusID:160025533}.

\bibitem[Scikit-Learn(2024)]{sklearn_sgdclassifier}
Scikit-Learn, D.
\newblock Sgdclassifier, 2024.
\newblock URL \url{https://scikit-learn.org/stable/modules/generated/sklearn.linear_model.SGDClassifier.html}.
\newblock Accessed: 2024-11-18.

\bibitem[Srivastava et~al.(2014)Srivastava, Hinton, Krizhevsky, Sutskever, and Salakhutdinov]{dropout}
Srivastava, N., Hinton, G., Krizhevsky, A., Sutskever, I., and Salakhutdinov, R.
\newblock Dropout: A simple way to prevent neural networks from overfitting.
\newblock \emph{Journal of Machine Learning Research}, 15\penalty0 (56):\penalty0 1929--1958, 2014.
\newblock URL \url{http://jmlr.org/papers/v15/srivastava14a.html}.

\bibitem[Tibshirani(1996)]{tibshirani1996}
Tibshirani, R.
\newblock Regression shrinkage and selection via the lasso.
\newblock \emph{Journal of the Royal Statistical Society: Series B (Statistical Methodology)}, 58\penalty0 (1):\penalty0 267--288, 1996.

\bibitem[Vaswani et~al.(2017)Vaswani, Shazeer, Parmar, Uszkoreit, Jones, Gomez, Kaiser, and Polosukhin]{attn}
Vaswani, A., Shazeer, N., Parmar, N., Uszkoreit, J., Jones, L., Gomez, A.~N., Kaiser, L., and Polosukhin, I.
\newblock Attention is all you need.
\newblock In \emph{Proceedings of the 31st International Conference on Neural Information Processing Systems}, NIPS'17, pp.\  6000–6010, Red Hook, NY, USA, 2017. Curran Associates Inc.
\newblock ISBN 9781510860964.

\end{thebibliography}
\bibliographystyle{icml2021}
\newpage

\onecolumn
\appendix
\csection{Appendix}

\csubsection{Analysis of the Original Work - GPT 2} \label{sec:gpt}
The original paper presents an idea for computing a basis expansion of the data using a generative pre-trained transformer (GPT). The primary idea behind this experiment was to construct a context-aware embedding of each vector. However, upon inspecting the source code we noticed several issues with this approach. Fundamentally, the work using GPT as a basis expansion tool is floored and shows a lack of understanding of the architecture of large language models (LLMs).

Given a sequence of tokens constructed from a sentence, GPT is trained to minimise the negative log-likelihood of the next token \cite{gpt}:
\[L_{1}(U) = -\sum_{i} \log{P(u_{i}|u_{i-k},...,u_{i-1};\Theta)}\] 
In this way, it seeks to complete sentences conditioned on the previous words. This is not to say that a GPT model cannot be used for other explicit tasks, this was explored in the original work, coined zero-shot or few-shot learning \cite{gpt2} \cite{gpt3}. Therefore, generating an embedding from a vector using a description is not necessarily a poor method. However, it is a strong deviation from the original task of GPT.

Regardless, there are several issues with how GPT was used as a basis expansion tool. Firstly, in the source code, the prompt asks the model to “Classify the following vector for a binary classification task”. This does not match the original instruction from the paper: “Please encode the following vector for a binary classification task”. These 2 tasks are very different, with the former asking to predict based on the vector, and the latter asking to encode based on the vector.

Secondly, the basis expansion is taken as the mean of all outputs from the model. However, given a sequence, GPT will output a vector of embeddings where each entry corresponds to the token used as the query, with the rest acting as keys/values to predict the token following the query \cite{Foster2019}[p239-252]. GPT uses a causal mask to prevent the latter keys in the sequence from being used to predict the token after the query \cite{attn} \cite{gpt}. Therefore, averaging the embeddings is not logical. Rather, the final embedding should be taken which would be used to predict the token immediately following the sentence, in this case, our basis expansion.

\begin{figure}[H]
    \begin{center}
        \includegraphics[width=0.6\textwidth]{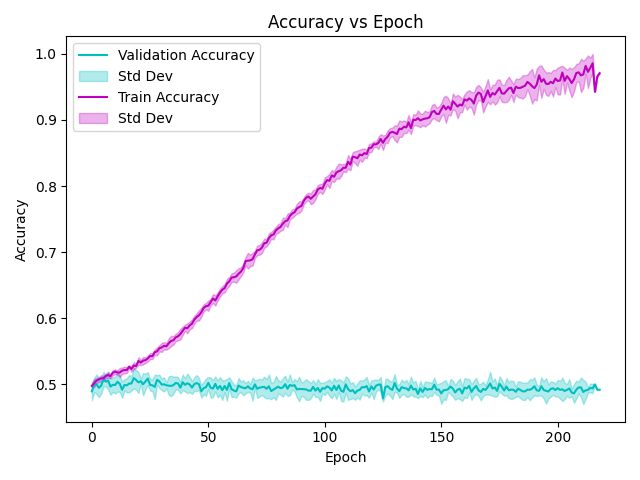}
        \caption{Network Trained on GPT2 Outputs}
        \label{fig:neural_net_gpt}
    \end{center}
\end{figure}

Lastly, generating an embedding for each vector separately means that the model will not have access to any other vectors as no fine-tuning occurs. It will not be able to produce an embedding accounting for the variation of features across all data points. This is essentially equivalent to generating noise for each data point and using this to train a model. In this case, we would expect a model to be able to fit the noisy training data well but generalize very poorly to unseen data. We support this hypothesis by training a neural network on the final embedding with the correct prompt. The graph of training is shown in figure \ref{fig:neural_net_gpt}.

\csubsection{Detailed Steps of Revealing the Dataset's Secret} \label{appendix:steps}
The discovery process leveraged iterative experimentation and observation of model behaviours across datasets with varying dimensions:
\begin{enumerate}
\itemsep0em 
    \item \textbf{Initial Observations}
    Logistic Regression struggled in $N=9$ with underfitting, when the polynomial expansion degree was low. This hinted the need for higher-order feature interactions to linearly separate the data.

    \item \textbf{Threshold Identification}
    Upon testing with a higher polynomial degree, performance improved significantly as the degree rose from 5 to 6, indicating that a sixth-degree interaction was sufficient to capture the necessary feature information for $N=9$.

    \item \textbf{Dimension Scaling Challenges}
    As $N$ increased, optimisation became slower due to the curse of dimensionality. Switching to a gradient-descent-based optimiser, i.e. \texttt{SGDClassifier} in \texttt{scikit-learn}, mitigated this issue.\cite{sklearn_sgdclassifier}

    \item \textbf{Overfitting in Higher Dimensions}
    For $N=15$, the model achieved near-perfect training accuracy (99\%) but failed on the validation and test sets (64\%), signalling overfitting with L2 regularization, which revealed evenly distributed low weights across coefficients.

    \item \textbf{[Key] Transition to L1 Regularization}
    Employing L1 regularization improved generalization. It reduced most weights to zero, leaving only one significant coefficient corresponding to a specific high-order feature interaction.\\
    \emph{Example discovery}: For $N=15$, the interaction $x_2 x_5 x_7 x_9 x_{10} x_{11} x_{12} x_{13} x_{14} x_{15}$ was identified as the key contributor to the dataset’s separability.

    \item \textbf{[Key] Feature Selection according to Distribution Shape}
    Due to the limitation of RAM storage space of lab machines, we weren't allowed to do an all-sweeping polynomial basis expansion when $N$ reaches 30. However, based on the \emph{informative feature} set we have already procured in N = 9, 12, 15, 18 datasets, we identified the following which was hardly a coincidence:
    \vspace{-1em}
    \begin{itemize}
        \itemsep0em 
        \item Every $N$ in the datasets is a multiple of 3.
        \item According to discovery in Section \ref{sec:data_exp} and figure \ref{fig:pmf} in Appendix \ref{appendix:pmf}, the Probability Mass Function (PMF) of features displays 3 shape patterns despite the number of $N$, which we will address as \emph{burst-like}, \emph{gaussian-like} and \emph{spread-out} later.
        \item The number of irrelevant features is always $1/3$ of the total feature number.
    \end{itemize}
    \vspace{-1em}
    After cross-referencing the irrelevant features we already extracted and the distribution shape of each feature, we discovered that features with \emph{burst-like} shapes perfectly matching the \emph{irrelevant feature} set. Thus, the selection of features was based on the PMF shape of each feature.

    \item \textbf{XOR conjecture}  
    During these investigations, we strongly suspected that the Kryptonite-N dataset might be modelled as a high-dimensional XOR problem. This intuition stemmed from the historical significance of the XOR problem in machine learning, which was once considered a challenging ``unsolvable problem" for simple models like single-layer perceptrons \cite{minsky1988perceptrons}. Given the context of this coursework and the apparent effort to design a dataset that would stump conventional approaches, it seemed plausible that the dataset's creator might intentionally incorporate XOR-like characteristics to test our limits.
    
    Although this was purely speculative, we decided it was worth exploring based on this ``smart trick" hypothesis. After \emph{selecting features} based on their PMF shape, we subsequently wrote out a \emph{discretizing function} to map the original feature to 0 and 1 based on their position in the distribution, and a \emph{xor function} to output final results. And it worked well.
\end{enumerate}

\emph{\textbf{The final reveal was nothing short of exhilarating, in a Kent-outsmarting-Luther and Wayne-besting-Quinn fashion.}}

\csubsection{Data Exploration Diagrams} \label{appendix:pmf}

\begin{figure}[H]
    \centering
    \includegraphics[width=0.45\textwidth]{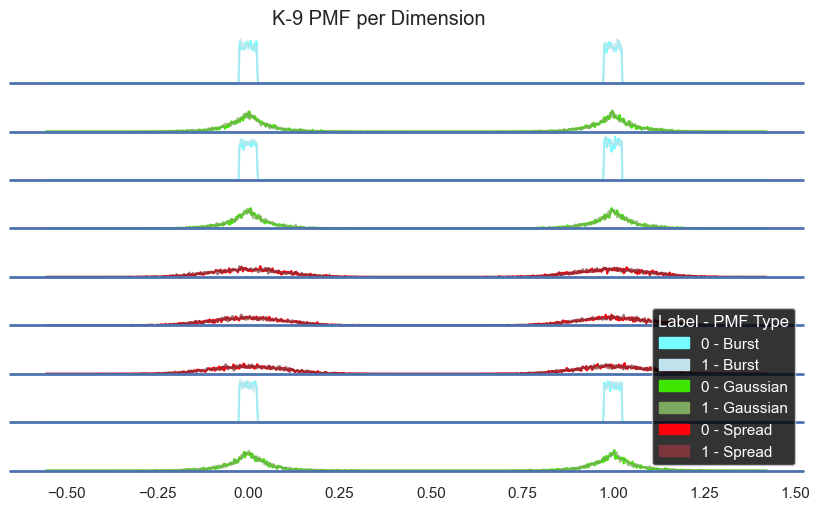}
    \caption{PMF Across Dimensions}
    \label{fig:pmf}
\end{figure}

\begin{figure}[H]
    \centering
    \includegraphics[width=0.4\textwidth]{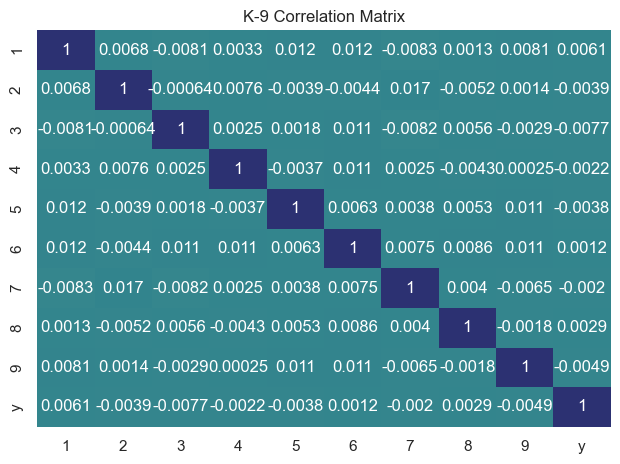}
    \caption{Correlation Matrix of Dimensions and Label}
    \label{fig:CorrMat}
\end{figure}

\begin{figure}[H]
    \centering
    \includegraphics[width=0.4\textwidth]{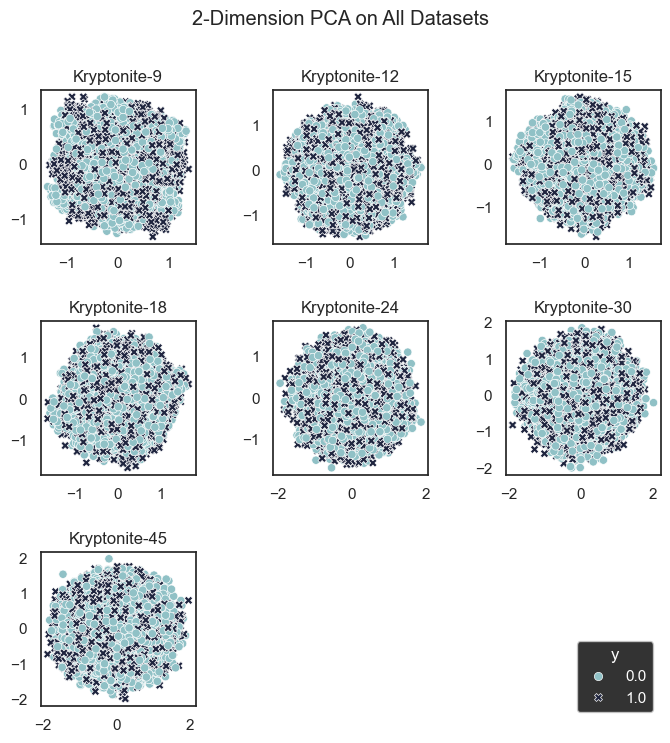}
    \caption{Correlation Matrix of Dimensions and Label}
    \label{fig:PCA}
\end{figure}

\begin{figure}[H]
    \centering
    \includegraphics[width=0.4\textwidth]{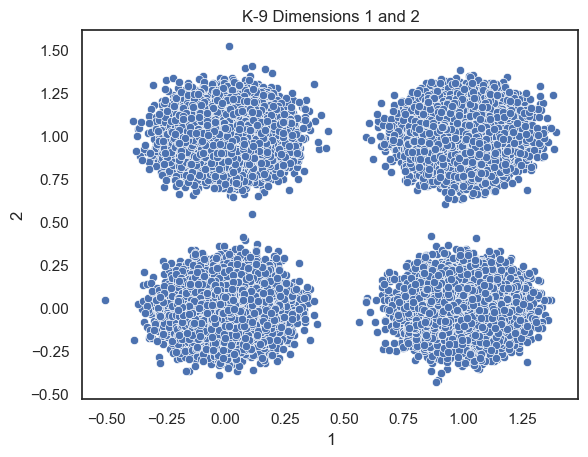}
    \caption{Dimensions 1 and 2}
    \label{fig:2Dim}
\end{figure}

\csubsection{Methodology Formulae}

\csubsubsection{Binary Cross Entropy Loss} \label{appendix:bce}

Loss functions are evaluated on the output of a neural network and used as a way to score how well the network fits the data. The task presented can be interpreted as a binary classification problem where each example \(x_{i}\) is classified as 0 or 1. A common loss function to use for this is the binary cross-entropy loss, which we show below. Where \(\hat{y_{i}}\) is the output from the network, and \(y_{i}\) is the ground truth. 

\[L_{BCE} = -\frac{1}{N}\sum_{i=1}^N(y_{i}log(\hat{y_{i}}) + (1-y_{i})log(1-\hat{y_{i}}))\]

\csubsubsection{Common Activation Functions} \label{appendix:act}
Amongst many activation functions, 3 were used in this work. The Sigmoid activation function has the formula \(\sigma(z) = \frac{1} {1 + e^{-z}}\), which is used to output a probability value. In our case, it was used to output the probability of the data sample belonging to the positive class. 

Tanh was used in the intermediate layers of our networks and is given by the formula: \(tanh(z) = \frac{e^z - e^{-z}}{e^z + e^{-z}} = \frac{1 - e^{-2z}}{1 + e^{-2z}}\).

Finally, the ELU activation function was also used in intermediate layers and is given by: \(ELU_{a}(z) = a(exp(z)-1) \text{ if } z < 0, z \text{ if } z \geq 0\). Note it introduces an additional hyperparameter \(a\).

\csubsubsection{SGD with Momentum Optimizer} \label{appendix:sgd}

\vspace{-0.3em}
\[m \leftarrow \beta m - \alpha\nabla_{\theta}L(\theta)\]
\[\theta \leftarrow \theta + m\]

\csubsubsection{Adam Optimizer} \label{appendix:adam}
\vspace{-2em}
\begin{gather*}
m \leftarrow \beta_{1} m - (1-\beta_{1})\nabla_{\theta}L(\theta)\\
s \leftarrow \beta_{2} s - (1-\beta_{2})\nabla_{\theta}L(\theta)^2\\
\hat{m} \leftarrow \frac{m}{1-\beta_{1}^t}\\
\hat{s} \leftarrow \frac{s}{1-\beta_{2}^t}\\
\theta \leftarrow \theta + \alpha\hat{m}\oslash\sqrt{\hat{s}+\epsilon}\\
\end{gather*}
\vspace{-2em}

\csubsection{Polynomial Basis Expansion}
\ref{appendix:poly_basis_plot}

\begin{figure}[H]
    \begin{center}
        \includegraphics[width=0.6\textwidth]{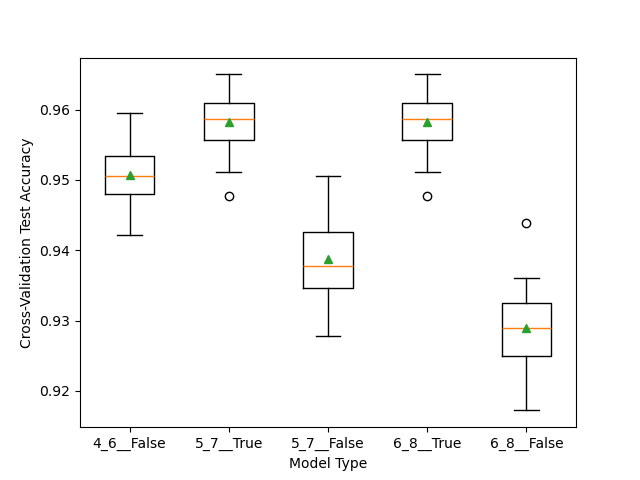}
        \caption{Cross Validation Accuracy Score Distribution of Logistic Regression Model with varying degrees of Polynomial basis}
        \label{fig:n_9_box1}
    \end{center}
\end{figure}

Figure \ref{fig:n_9_box1} displays cross-validation test accuracy scores over the varying ranges of polynomial degrees for N=9. Model Type is labelled by the degree range used (min\_max) followed by whether or not \textit{interaction\_only} is set to True or False.

\csubsection{Neural Network Architectures}

\begin{table}[H]
\caption{Neural Network Architectures for Each Dataset}
\label{table:nn}
\vskip 0.15in
\begin{center}
\begin{small}
\begin{sc}
\begin{tabular}{llll}
\toprule
\multicolumn{3}{l}{Neural Network Architectures} \\
\midrule
Dataset    & Optimiser   & Activation  \\
N-9        & SGD w/ Momentum  & ELU \\
N-12       & SGD w/ Momentum &  ELU               \\
N-15       & SGD w/ Momentum &  Tanh              \\
N-18       & Adam &  Tanh               \\
\bottomrule
Learning Rate & Dropout & Batch Size   \\
0.01 & None & 128 \\
0.1 & 0.01 & 256 \\
0.1 & 0.03 & 256 \\
0.1 & 0.07 & 256 \\
\bottomrule
\end{tabular}
\end{sc}
\end{small}
\end{center}
\vskip -0.1in
\end{table}

\csubsection{Neural Network Training Graphs}

\begin{figure}[H]
    \begin{center}
        \includegraphics[width=0.6\textwidth]{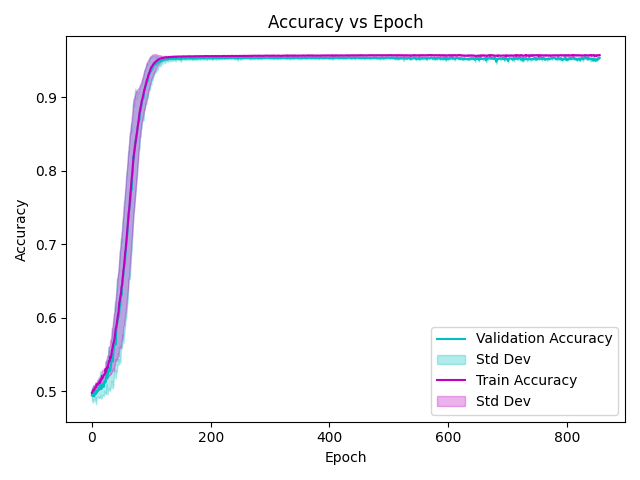}
        \caption{Training vs Validation Loss for Neural Network Trained on N9 Dataset}
        \label{fig:neural_net_results_N9}
    \end{center}
\end{figure}

\begin{figure}[H]
    \begin{center}
        \includegraphics[width=0.6\textwidth]{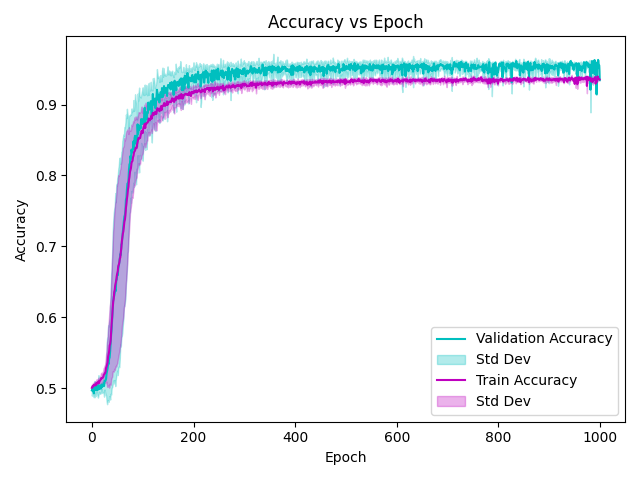}
        \caption{Training vs Validation Loss for Neural Network Trained on N12 Dataset}
        \label{fig:neural_net_results_N12}
    \end{center}
\end{figure}

\begin{figure}[H]
    \begin{center}
        \includegraphics[width=0.6\textwidth]{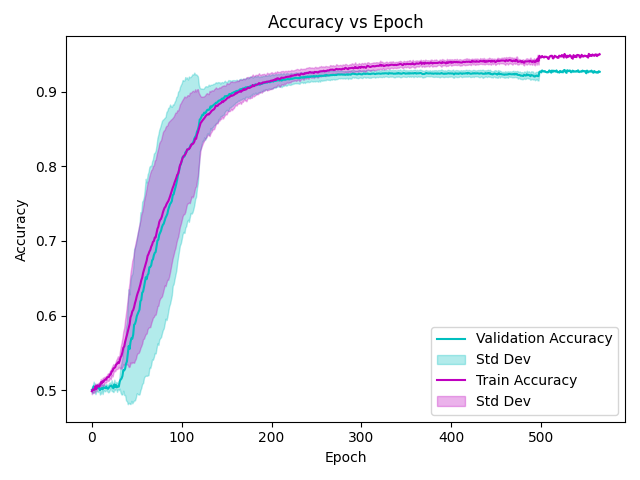}
        \caption{Training vs Validation Loss for Neural Network Trained on N15 Dataset}
        \label{fig:neural_net_results_N15}
    \end{center}
\end{figure}

\begin{figure}[H]
    \begin{center}
        \includegraphics[width=0.6\textwidth]{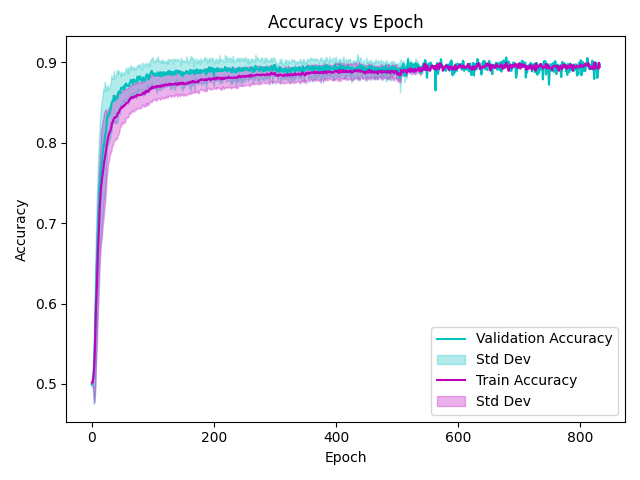}
        \caption{Training vs Validation Loss for Neural Network Trained on N18 Dataset}
        \label{fig:neural_net_results_N18}
    \end{center}
\end{figure}

Figures \ref{fig:neural_net_results_N9} to \ref{fig:neural_net_results_N18} show the training and validation losses for the final tuned neural networks trained on datasets N9-N18.


\csubsection{Neural Network Overfitting Comparison}

\begin{figure}[H]
    \begin{center}
        \includegraphics[width=0.6\textwidth]{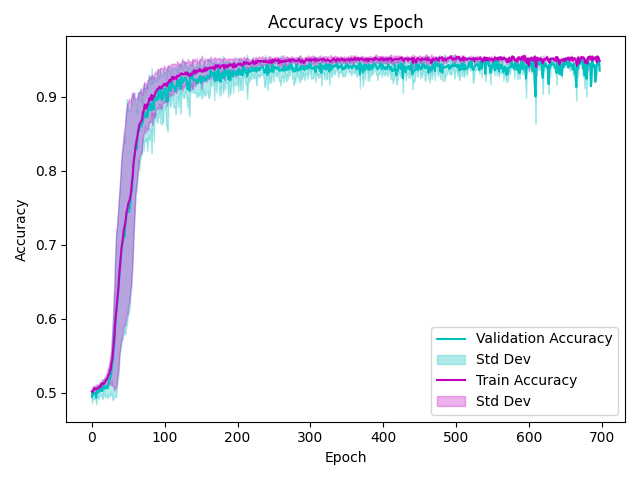}
        \caption{Neural Network Trained on N12 Dataset with no Dropout}
        \label{fig:overfit_N12}
    \end{center}
\end{figure}

Figure \ref{fig:overfit_N12} shows the results of training the neural network on the N12 dataset with no dropout. As can be seen, when compared to figure \ref{fig:neural_net_results_N12}, the network shows more overfitting on the training data.

\begin{figure}[H]
    \begin{center}
        \includegraphics[width=0.6\textwidth]{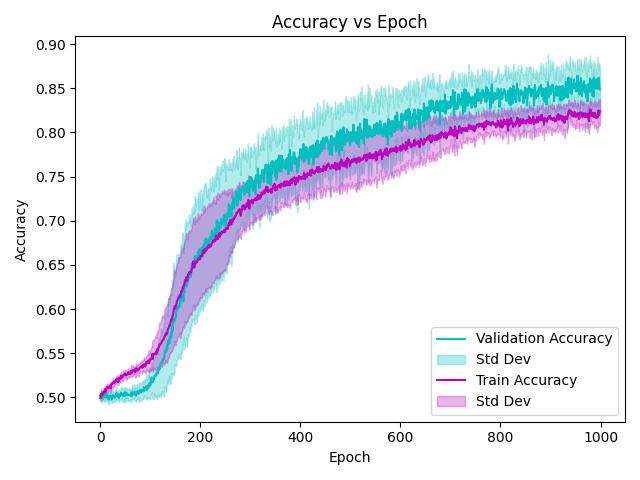}
        \caption{Neural Network Trained on N15 Dataset with ELU Activation Layer}
        \label{fig:overfit_N15_1}
    \end{center}
\end{figure}

Figure \ref{fig:overfit_N15_1} shows the results of training the neural network on the N15 dataset with an ELU activation function rather than a Tanh activation function.

\begin{figure}[H]
    \begin{center}
        \includegraphics[width=0.6\textwidth]{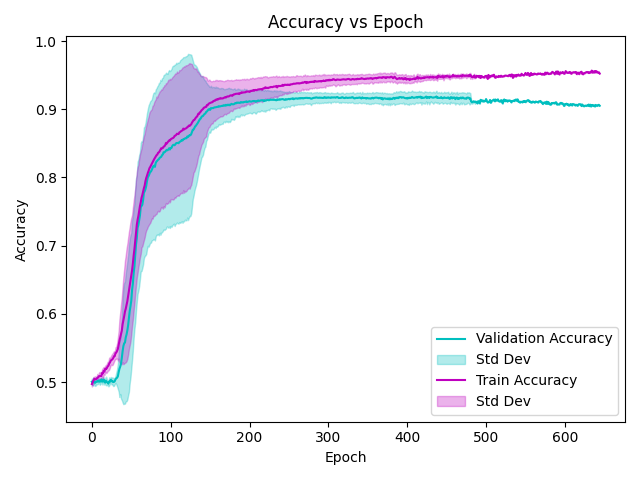}
        \caption{Neural Network Trained on N15 Dataset with 0.01 Dropout}
        \label{fig:overfit_N15_2}
    \end{center}
\end{figure}

Figure \ref{fig:overfit_N15_2} shows the results of training the neural network on the N15 dataset with 0.01 dropout compared to the original 0.03 dropout. As can be seen, when compared to figure \ref{fig:neural_net_results_N15}, the network shows more overfitting on the training data.

\begin{figure}[H]
    \begin{center}
        \includegraphics[width=0.6\textwidth]{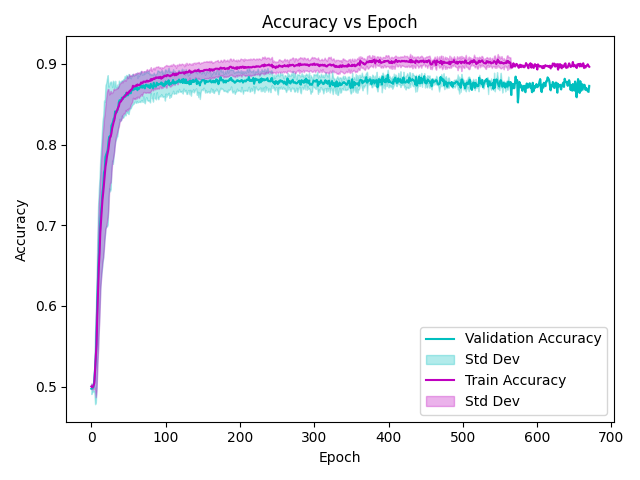}
        \caption{Neural Network Trained on N18 Dataset with 0.03 Dropout}
        \label{fig:overfit_N18}
    \end{center}
\end{figure}

Figure \ref{fig:overfit_N18} shows the results of training the neural network on the N18 dataset with 0.03 dropout compared to the original 0.07 dropout. As can be seen, when compared to figure \ref{fig:neural_net_results_N18}, the network shows more overfitting on the training data.

\csubsection{Results on GPT}

Figure \ref{fig:neural_net_gpt} shows the results for training a neural network on feature vectors outputted from a neural network. As can be seen, the model produces strong training accuracy, but greatly overfits the data, producing poor generalization results.

\csubsection{Sustainability Analysis Graphs} \label{appendix:carbon}

\begin{figure}[H]
    \centering
    \subfigure[Training Emissions]{\includegraphics[width=0.45\textwidth]{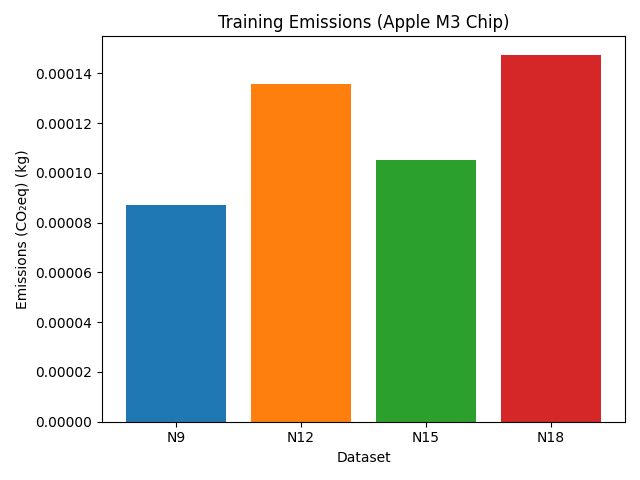}} 
    \subfigure[Inference Emissions]{\includegraphics[width=0.45\textwidth]{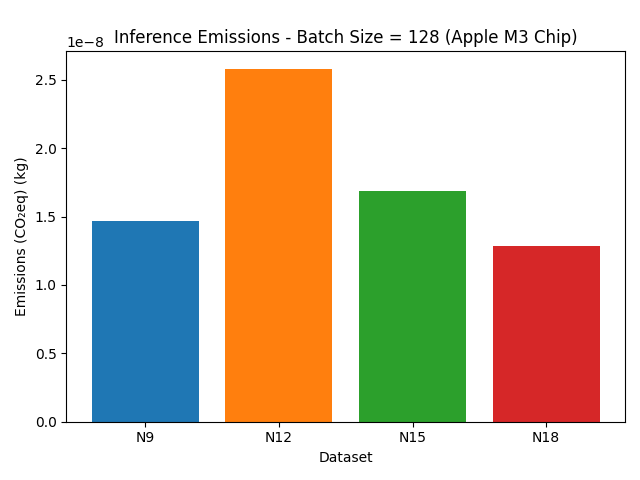}} 
    \caption{Emissions Produced for Training and Inference for an Apple M3 Chip}
    \label{fig:carbon_impact}
\end{figure}

Figure \ref{fig:carbon_impact} shows the difference in emissions produced for training and inference across datasets on an Apple M3 system.

\begin{figure}[H]
    \centering
    \subfigure[]{\includegraphics[width=0.4\textwidth]{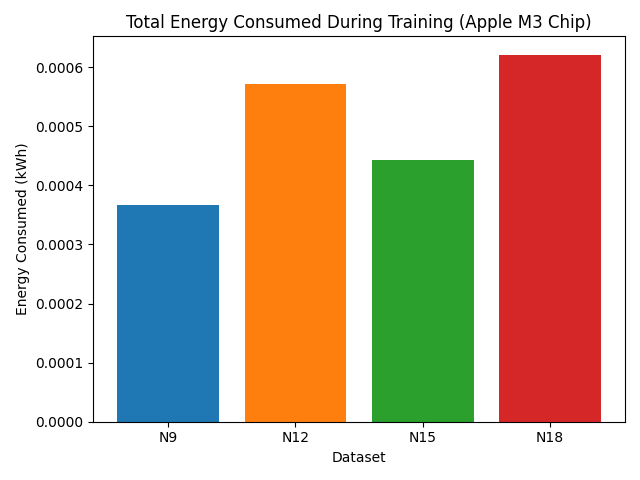}} 
    \subfigure[]{\includegraphics[width=0.4\textwidth]{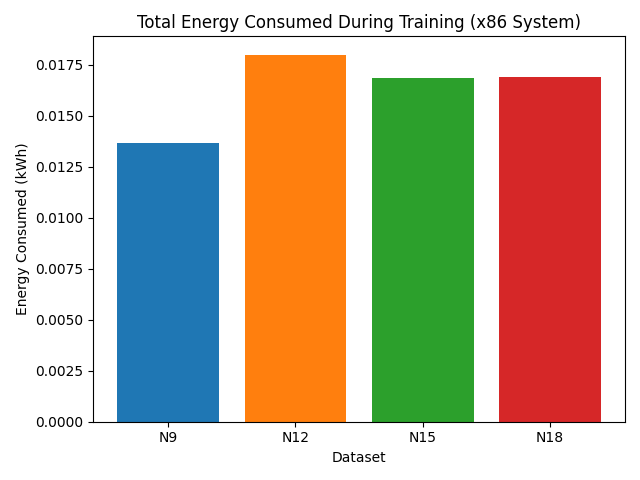}} 
    \caption{Energy Consumed During Training for an Apple M3 System vs x86 system (AMD Ryzen 7 5700X)}
    \label{fig:energy_impact_comp_train}
\end{figure}

Figure \ref{fig:energy_impact_comp_train} shows the difference in energy consumption during training for an Apple M3 system vs an AMD Ryzen 7 5700X (x86) system. 

\begin{figure}[H]
    \centering
    \subfigure[]{\includegraphics[width=0.48\textwidth]{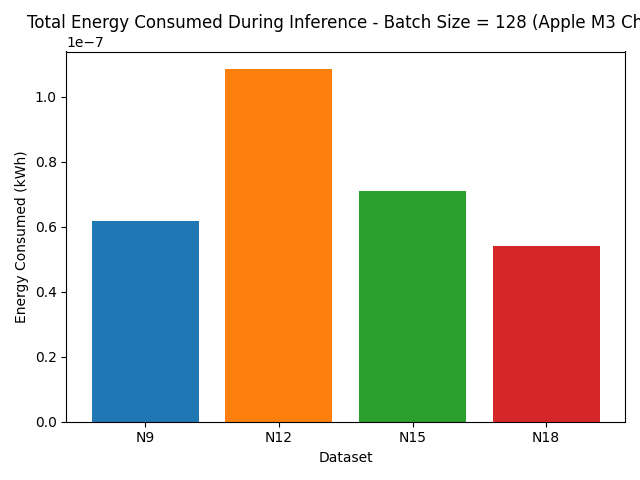}} 
    \subfigure[]{\includegraphics[width=0.48\textwidth]{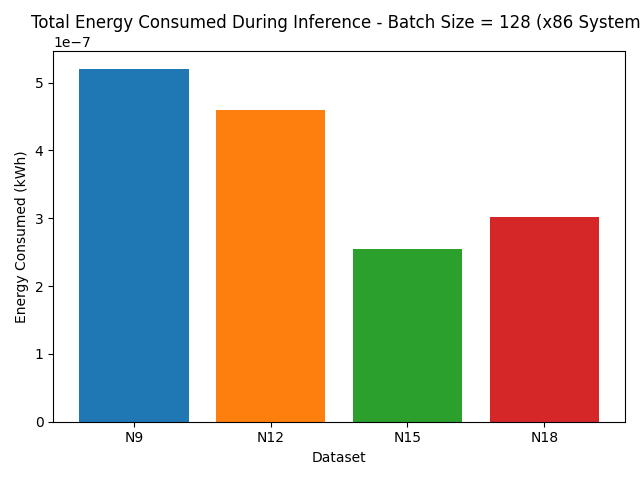}} 
    \caption{Energy Consumed During Inference for an Apple M3 System vs x86 system (AMD Ryzen 7 5700X)}
    \label{fig:energy_impact_comp_test}
\end{figure}

Figure \ref{fig:energy_impact_comp_test} shows the difference in energy consumption during inference for an Apple M3 system vs an AMD Ryzen 7 5700X (x86) system. 

\begin{figure}[H]
    \centering
    \subfigure[]{\includegraphics[width=0.48\textwidth]{figures/carbon_impact/training_emissions.png}} 
    \subfigure[]{\includegraphics[width=0.48\textwidth]{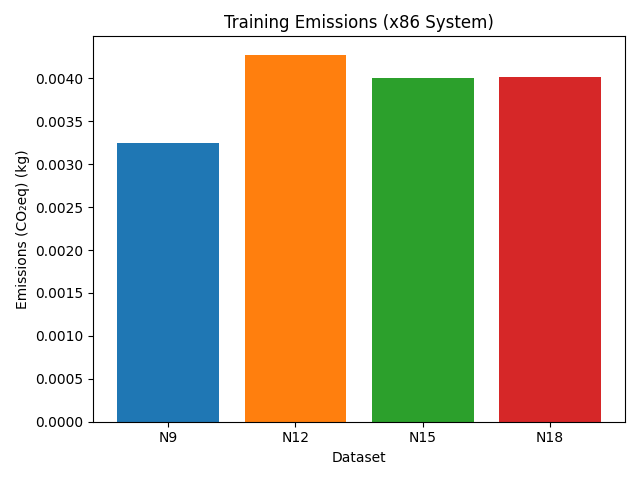}} 
    \caption{Emissions Produced During Training for an Apple M3 System vs x86 system (AMD Ryzen 7 5700X)}
    \label{fig:carbon_impact_comp_train}
\end{figure}

Figure \ref{fig:carbon_impact_comp_train} shows the difference in emissions during training for an Apple M3 system vs an AMD Ryzen 7 5700X (x86) system. 

\begin{figure}[H]
    \centering
    \subfigure[]{\includegraphics[width=0.48\textwidth]{figures/carbon_impact/test_emissions.png}} 
    \subfigure[]{\includegraphics[width=0.48\textwidth]{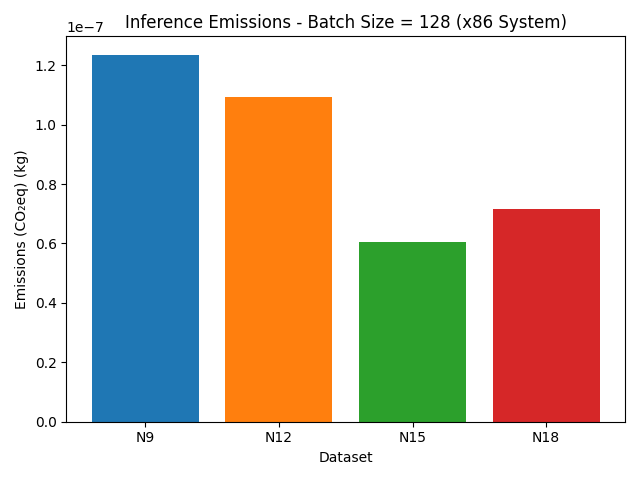}} 
    \caption{Emissions Produced During Inference for an Apple M3 System vs x86 system (AMD Ryzen 7 5700X)}
    \label{fig:carbon_impact_comp_test}
\end{figure}

Figure \ref{fig:carbon_impact_comp_test} shows the difference in emissions during inference for an Apple M3 system vs an AMD Ryzen 7 5700X (x86) system. 

\end{document}